\documentclass[letterpaper, 10 pt, conference]{ieeeconf}
\IEEEoverridecommandlockouts

\usepackage{cite}
\usepackage{amsmath,amssymb,amsfonts}
\usepackage{algorithmic}
\usepackage{textcomp}
\usepackage{stmaryrd}

\usepackage{xcolor}
\def\BibTeX{{\rm B\kern-.05em{\sc i\kern-.025em b}\kern-.08em
    T\kern-.1667em\lower.7ex\hbox{E}\kern-.125emX}}

\usepackage{epsfig}
\usepackage{graphicx}
\usepackage{multirow}
\usepackage{tabularx,booktabs}
\usepackage{todonotes}
\usepackage{caption}
\usepackage{subcaption}
\DeclareMathOperator*{\argmin}{arg\,min}
\title{
Automatic Image Annotation for Mapped Features Detection 
}

\author{
Maxime Noizet$^1$
\and
Philippe Xu$^{1,2}$
\and
Philippe Bonnifait$^1$
\thanks{$^{1}$The authors are with the Universit\'e de Technologie de Compi\`egne, CNRS, Heudiasyc, France. $^{2}$The author is with ENSTA Paris, Institut Polytechnique de Paris, U2IS, Palaiseau, France. This work has been funded by the European project ERASMO~ (GSA/GRANT/03/2018) in the framework of the SIVALab laboratory between Renault and Heudiasyc.}
}

\begin{document}

\maketitle
\thispagestyle{empty}
\pagestyle{empty}

\begin{abstract}

Detecting road features is a key enabler for autonomous driving and localization. For instance, a reliable detection of poles which are widespread in road environments can improve localization.
Modern deep learning-based perception systems need a significant amount of annotated data. Automatic annotation avoids time-consuming and costly manual annotation. Because automatic methods are prone to errors, managing annotation uncertainty is crucial to ensure a proper learning process. 
Fusing multiple annotation sources on the same dataset can be an efficient way to reduce the errors.
This not only improves the quality of annotations, but also improves the learning of perception models.  
In this paper, we consider the fusion of three automatic annotation methods in images: feature projection from a high accuracy vector map combined with a lidar, image segmentation and lidar segmentation.
Our experimental results demonstrate the significant benefits of multi-modal automatic annotation for pole detection through a comparative evaluation on manually annotated images.
Finally, the resulting multi-modal fusion is used to fine-tune an object detection model for pole base detection using unlabeled data, showing overall improvements achieved by enhancing network specialization. The dataset is publicly available. 

\end{abstract}


\section{Introduction}

\begin{figure}
    \centering
    \includegraphics[width=\columnwidth]{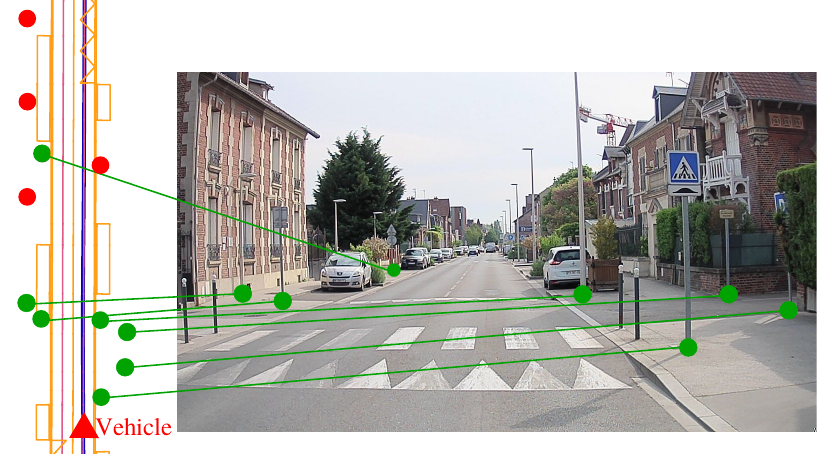}
    \caption{Illustration of the connections between georeferenced  poles on the map (left) and an image captured by a vehicle (right). The pole bases in red are not visible from the camera. Note that the map does not include the four black bollards on the sides of the crosswalk.}
    \label{fig:hd_map}
\end{figure}

Localization is a core functionality for autonomous driving.
Depending on the navigation context, obtaining a reliable and accurate localization only through GNSS and dead-reckoning sensors can be challenging. 
A complementary solution is integrating maps with exteroceptive sensors, such as cameras or lidars, capable of measuring distances and/or angles relative to georeferenced map features.
In this paper, we consider that the map is given to the system beforehand by a map provider.
This map is agnostic to the perception sensors set up on the vehicle.
Therefore, contrary to simultaneous localization and mapping contexts where the features encoded in the map are directly linked to the ones detected by the sensors, one needs to build a perception module able to explicitly detect these map features as shown in Fig.~\ref{fig:hd_map}.

A standard vector map used for autonomous driving includes features of the road environment, such as lane markings, but also road furniture such as traffic signs, traffic lights or streetlamps.
These pole-like structures are widespread and represent optimal localization features.
In a 2D vector map, these features are commonly depicted as points encoding the positions of their bases on the ground.

Because the shape of a pole is geometrically constrained, lidars, which provide 3D geometric information, appear to be quite suitable for detecting poles. However, this proves difficult  due to data sparsity, especially when the poles are far away. 
Cameras, on the other hand, are capable of detecting distant poles, and offer the advantage of developing less expensive systems. 

The goal of this paper is to build a monocular vision module able to detect exclusively the poles stored in the map.
Note that some other road furniture such as bollards, which are also some types of poles, are not considered in the map as these structures are more prone to damage by road users and are less reliable in a longer period of times.
The perception system should therefore be able to discriminate mapped poles from others.


Contrary to classical work on object detection, we do not wish to build a generic pole detector. Indeed, detecting all the poles including some types of poles that are not present in the map may lead to a poorer localization performance because of more ambiguity in data association. Instead, we wish to learn an ad hoc detector linked to the map. This implies that we need some annotated training images of the area covered by the map which amounts to a signification workforce if done manually. Therefore, we propose an automatic pipeline that takes as input a map and some raw unlabeled data recorded in the area covered by the map, annotates the pole bases automatically in the images and train an appropriate detector dedicated to the given map.

Automatic annotation methods are inevitably prone to errors, introducing false annotations or missing poles in the environment, thus consequently limiting the detection performance. 
Special attention must be given to evaluating annotations and minimizing errors to provide robust detectors capable to accurately characterize pole bases. The approach taken in this article involves combining multiple sources of annotations to reduce errors.

In Section \ref{sec:sota}, we introduce some related works. In Section \ref{sec:annotation}, we present the fusion of annotation sources.  In Section \ref{sec:case_study}, we focus on a case study using map data, lidar and images to train a pole detector and a method for managing uncertain annotations. 
Finally, experimental results are detailed in Section \ref{sec:exp}, to describe first the annotation performance and then the performance of the detector. 


\section{Related works}
\label{sec:sota}

Pole-like features are commonly used for localization ~\cite{li_robust_2021,sefati_improving_2017,spangenberg_pole-based_2016,ITSC23}. Using geometric assumptions, lidar sensors are classically used to detect pole features\cite{gouda_fully_2022,lehtomaki_detection_2010,rodriguez-cuenca_automatic_2015}.
When it comes to detection tasks using cameras, deep learning techniques are generally employed. 
Poles are often detected through semantic segmentation at pixel-level, a method that not only requires pixel-level annotation but is also far less computationally efficient compared to bounding box based object detection methods such as YOLO \cite{yolov7}.

Training deep neural networks requires a significant amount of annotated data which becomes costly when made manually.
When no annotated data is available for a given task or context, it is interesting to provide an automatic annotation.
However, this kind of approach is prone to errors and can lead to lower performance compared to annotation made by humans.
Some automatic and semi-automatic annotation methods to build datasets have been proposed in \cite{dong23,Sun20,Lee21, yu2016lsun,Waymo3DAutoLabelling} providing hard labels, to be used in classical supervised training pipelines. 
In particular, pseudo-labeling, a process that involves annotating unlabeled data by using predictions obtained from a pre-trained network for subsequent retraining, is generally applied\cite{pseudo_label,sohn2020simple,9577846,data_distillation}. 
However, the pre-trained network must be consistent with the detection task at hand and pseudo-labeling can be insufficient or lead to poor results due to the low quality of labels.

The use of these approaches can potentially lead to significant annotation errors if the quality of the automatic annotations is not properly controlled which will decrease the performance of the learned model. 
Even though deep learning models may exhibit a small degree of tolerance for annotation errors in object detection contexts, the accumulation of numerous errors can significantly degrade the network performance\cite{chachula2023combating,8814137}.
Particular attention should therefore be given to the quality assessment of annotations.

To account for annotation errors, one can try to further clean out the errors, or instead manage and quantify uncertainties of the annotations. 
For instance, confident learning involves characterizing label quality using the model to prune label errors from training sets \cite{chachula2023combating,northcutt2021confident}. 
However, since the model is employed to identify errors, it seems challenging to prune labels corresponding to false positives that are similar to the objects we aim to detect. It is typically true with bollards we do not want to detect in our case. 

Some methods involve modifying the loss used in the network to handle uncertainties\cite{reed2014training} and manipulating soft-labels. However, modifying the loss or the network itself limits the usability of these approaches, potentially requiring more challenging work than selecting a state-of-the-art model to learn. Additionally, estimating uncertainties is not particularly straightforward and depends on the annotation approaches used.

That is why we want to design automatic annotation methods that provide accurate hard labels with the minimum possible errors.
To get good annotations automatically, multi-source approaches can be employed. For instance, in \cite{MS3D++}, the authors employed a set of pre-trained multi-frame 3D detectors in lidar point clouds and fuse their detections to build new pseudo-labels on an unlabeled set.


\section{Multi-modal annotation}
\label{sec:annotation}
\subsection{Problem statement} 

Consider the problem of the fusion of image annotations computed from different sources in order to reach a better set of annotations for learning. Throughout this work, we only address the problem of detecting a single type of object. 

For a given method \(k\), the set of annotations for an image \(i\) is defined as
\begin{align}
    {}^{(k)}a^i = \left\{ \left. {}^{(k)}a^i_j \right| j=1,\ldots, {}^{(k)}n^i\right\},
\end{align}
where \({}^{(k)}n^i\) is the number of detected objects in image~\(i\) by the method \(k\) and \({}^{(k)}a^i_j\) encodes an object annotation, which is here the coordinates of a single point.

For \(N\) images, the resulting annotation set by method \(k\) is
\begin{align}
    {}^{(k)}a = \left\{ \left. {}^{(k)}a^i \right| i=1,\ldots, N \right\}.
\end{align}

Because each automatic annotation method is prone to errors (\textsl{i.e.} false positives and false negatives), the aim is to combine the annotations into a new set of annotations with a higher quality in terms of precision and/or recall.

\begin{figure*}[]
    \centering
    \includegraphics[width=\linewidth]{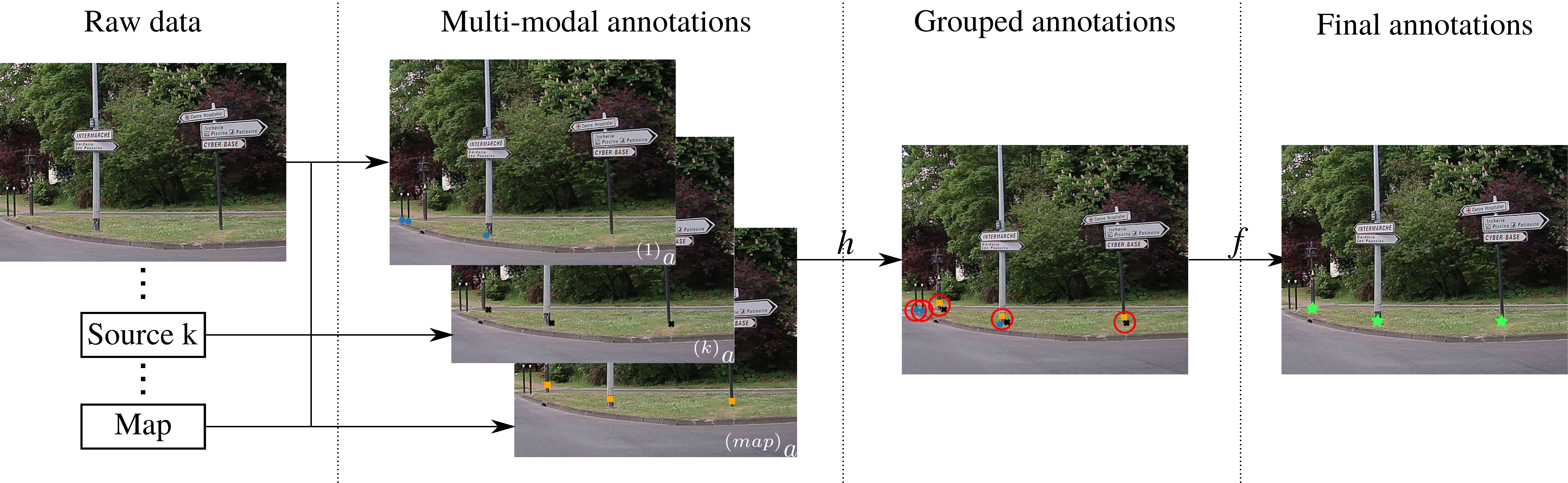}
    \caption{Steps in an automatic multi-modal labeling method.  Multiple annotation sets are obtained from the images and diverse data sources, including the vector map. Thanks to a data association function  \(h\), annotations are grouped to derive final annotations through a fusion function \(f\). The final annotations are displayed with green stars on the last image.}
    \label{fig:overview}
\end{figure*}

An overview of the entire annotation process is visible in Fig. \ref{fig:overview}. 
Annotations obtained through diverse methods are associated and then combined in order to generate a final annotation that should be as close as possible to the expected reference one.


\subsection{Fusion of annotations}
\label{sec:consensus}

The fusion process of \(K\) annotation sets coming from \(K\) independent methods can be decomposed into two principal steps.
The first step is to define a data association function \(h\) that, given a set of annotations \(A=\left\{ {}^{(1)}a^i, {}^{(2)}a^i, \ldots, {}^{(K)}a^i \right\}\), returns a set of clusters of annotations corresponding to the annotation of the same element along different modalities:
\begin{align}
    h(A) = \left\{ c^i_1, c^i_2,\ldots, c^i_M \right\}
\end{align}
where $M$ is the number of different annotated elements and each \(c^i_j\) is a set that contains at most one element from each \({}^{(k)}a^i\): 
\(c^i_j = \left\{ {}^{(k)}a^i_{\ell} \right\}\) if a pole is detected only by method \(k\), 
\(c^i_j = \left\{ {}^{(k)}a^i_{\ell}, {}^{(k')}a^i_{\ell'} \right\}\) if detected by methods \(k\) and \(k'\), and so on. 
In the case where all methods detect the pole, 
\(c^i_j = \left\{ {}^{(1)}a^i_{\ell_1}, \ldots, {}^{(K)}a^i_{\ell_K} \right\}\). Note that \(h(A)\) is a partition of the sets in \(A\):
\begin{equation}
    \forall j\neq\ell,\ c^i_j \cap c^i_{\ell} =\emptyset, \quad\text{and } \bigcup_{j=1,\ldots,M} c^i_j = \bigcup_{k=1,\ldots,K} {}^{(k)}a^i
\end{equation}

Since there are no explicit semantic classes encoded in the annotations, the criteria used in \(h\) is based on geometric proximity metrics, \textsl{e.g.}, Euclidean distance between points.

The second step is to create a fusion function \(f\) that combines annotations from a cluster \(c^i_j\) into one. This means determining the specific location in the image for the corresponding object. There are various approaches available. One can calculate the average point of annotations, or choose the best one based on a quality or confidence metric. 

From these two functions \(f\) and \(h\), we can define, for image \(i\), some consensus annotation sets defined as
\begin{align}
    {}^{(1:K)}_{\phantom{1111}q} a^i = \left\{ f(C) \left| C\in h(A) \text{ and } |C| \geq q \right. \right\}
    \label{eq:voting}
\end{align}
where \(|C|\) is the cardinality of the set of annotations \(C\) and \(q\in\{1,\ldots,K\}\) is a degree of consensus.
The set \({}^{(1:K)}_{\phantom{1111}q} a^i\) corresponds to the fused annotations of all the objects that have been annotated by at least \(q\) methods among the \(K\) ones.
Two particular sets \({}^{(1:K)}_{\phantom{1111}1} a^i\) and \({}^{(1:K)}_{\phantom{111}K} a^i\) corresponds to the union and the intersection of the annotations sets, respectively.
In other words, if an annotation belongs to \({}^{(1:K)}_{\phantom{1111}1} a^i\), it means that at least one method agrees with it while if it belongs to \({}^{(1:K)}_{\phantom{111}K} a^i\) then all the \(K\) methods agree.

Depending on the fusion strategy, an increase of precision or recall of the automatic annotation is expected. 
By applying the union of annotations, the resulting recall is guaranteed to increase since it provides more annotations. 
However, a decrease in precision may occur if false negatives are added.
The intersection of annotations can lead to a precision improvement, though not guaranteed, but inevitably decreases the recall since it removes some annotations.
The balance between precision and recall performance depends on the application, one can be favored over the other and conversely.

\section{Case study}
\label{sec:case_study}

We consider three annotation methods and their combinations to train a pole base detector. In this context, each annotation \( {}^{(k)}a^i_j\) corresponds to the coordinates  \((u_j,v_j)\) of the pole base in the image.

\begin{figure*}[]
    \centering
    \begin{subfigure}[t]{0.325\textwidth}
        \includegraphics[width=\columnwidth]{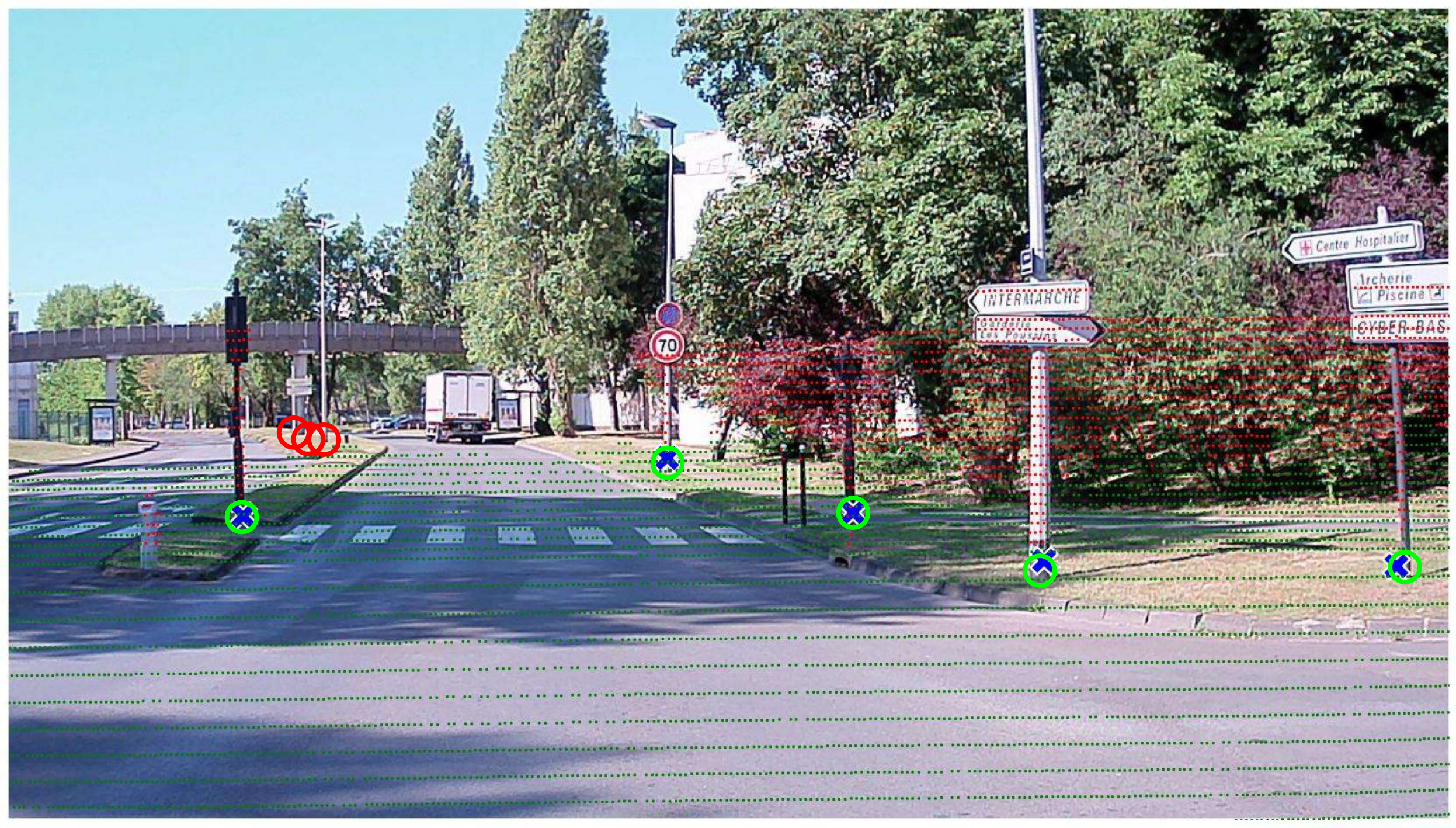}
        \caption{$M$ annotation using the map with lidar (point-cloud in green for ground).}
        \label{fig:map_annotation}
    \end{subfigure}
    \begin{subfigure}[t]{0.325\textwidth}
        \includegraphics[width=\columnwidth]{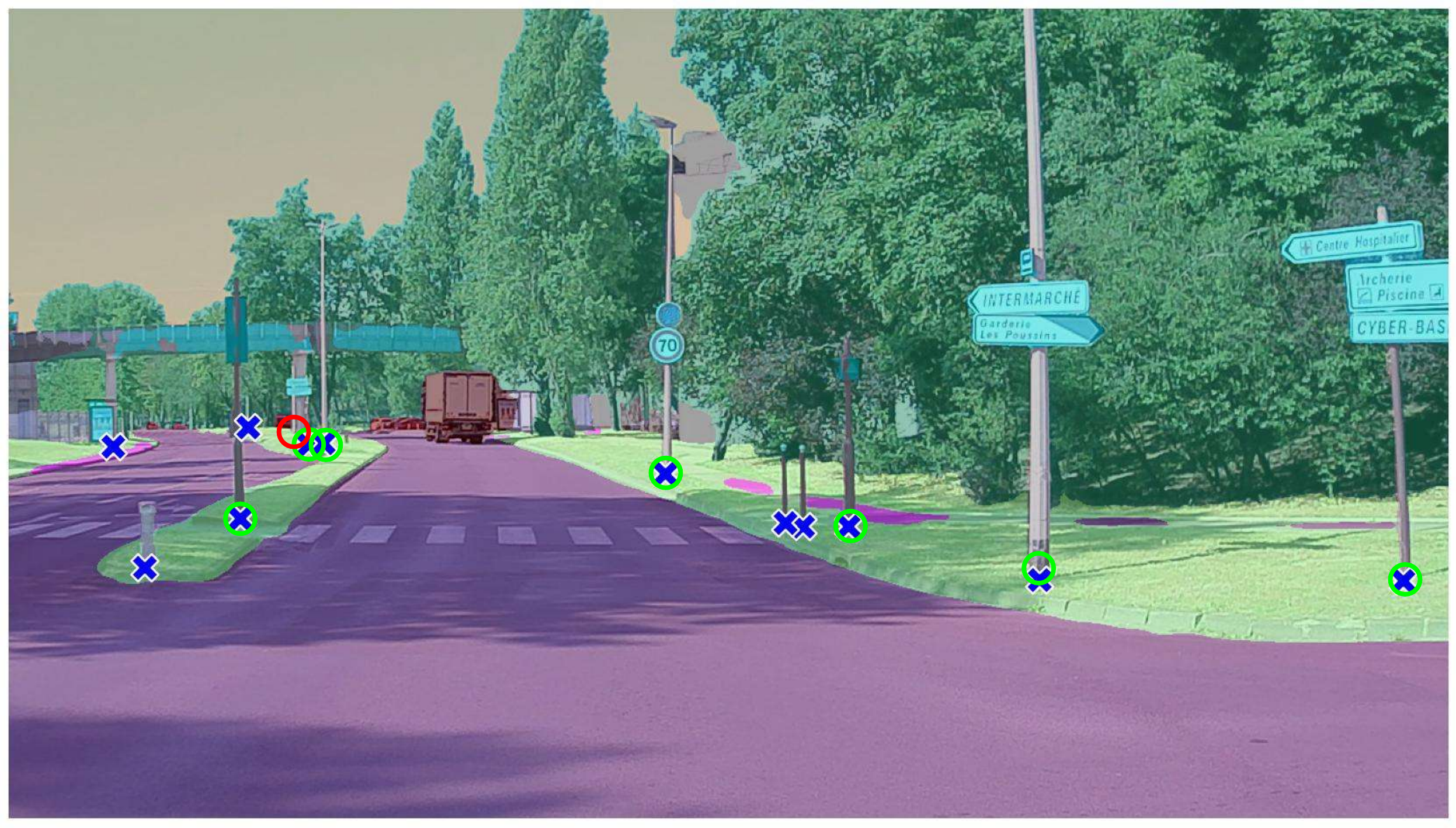}
        \caption{$S$ annotation overlaid with the segmentation mask.}
        \label{fig:segmentation_annotation}
    \end{subfigure}
    \begin{subfigure}[t]{0.325\textwidth}
        \includegraphics[width=\columnwidth]{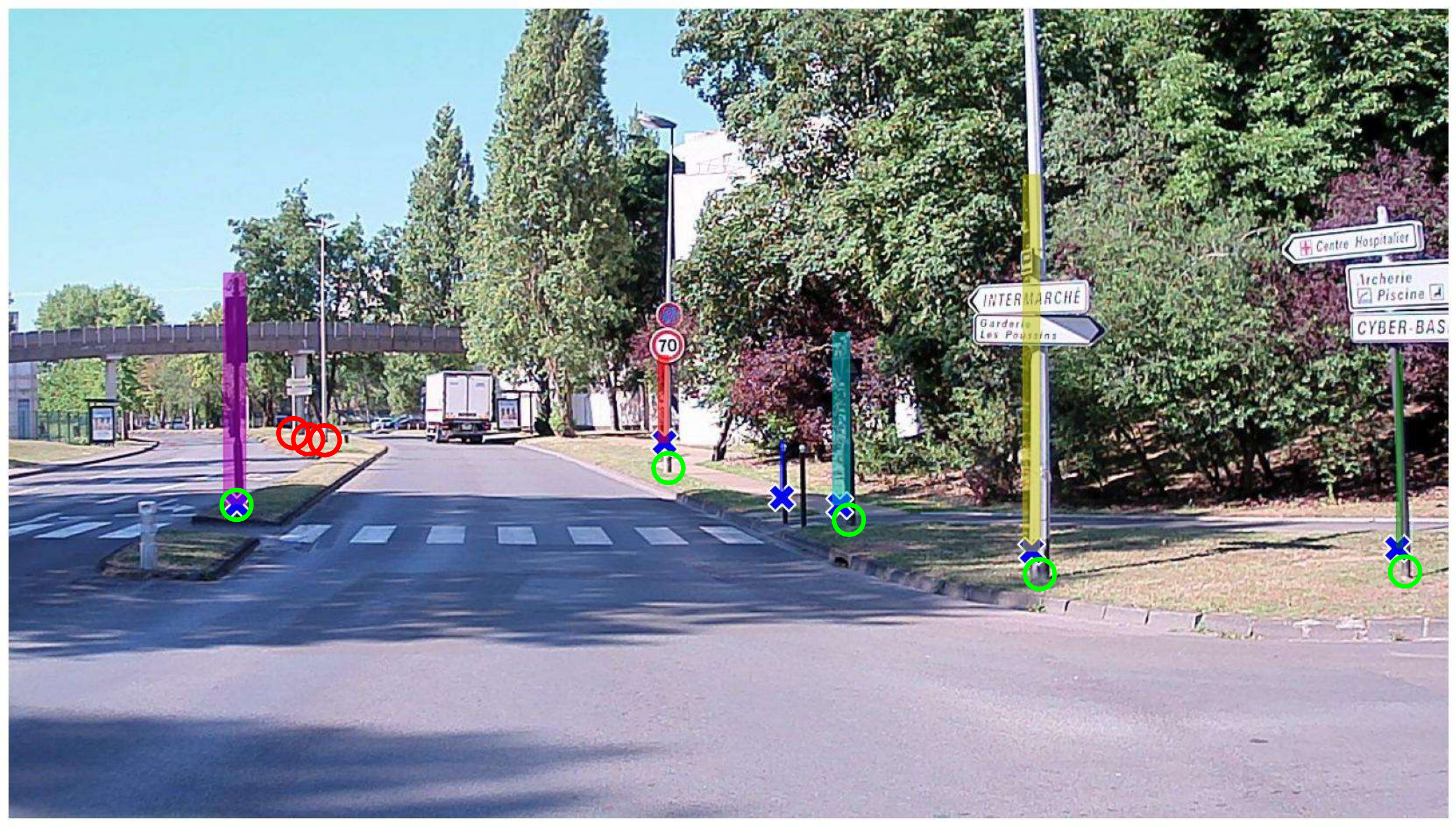}
        \caption{$L$ annotation with projected lidar clusters.}
        \label{fig:lidar_annotation}
    \end{subfigure}
    \caption{Examples of automatic annotations obtained using three different methods. They are depicted with blue crosses.
    Green circles represent reference annotations defined by humans and correctly annotated automatically.
    The red ones are those that are missed.
    }
    \label{fig:annotation}
\end{figure*}

\subsection{Pole base annotation methods}

\subsubsection{Map-based annotation (denoted M)} 
A first approach is to use a 2D High Definition (HD) vector map along with a high accuracy localization system. 
The poles georeferenced within the field of view of the camera are extracted from the map and projected onto the images.
Even though the poles contained in the map concern furniture (such as traffic signs, traffic lights or streetlamps) which is stable over time, the map can still become outdated. 
So, false positives and negatives are inevitable.
To project the 2D map features at the ground level, a lidar is used to estimate the ground and check for occluded pole bases. 
In this work, we use the ground segmentation method proposed in \cite{ground_seg} and the lidar pipeline as described in \cite{IV23}. 
Fig.~\ref{fig:map_annotation} illustrates some automatic annotations using this method. 
The blue crosses depict the generated annotations, the ones circled in green are correct annotations while the red circles correspond to missed poles.
In this particular example, the three missed poles are too far away from the vehicle making annotation impossible. This limitation is due to the lack of ground points, represented by green dots, near their bases, making it challenging to project the map points accurately. To mitigate the risk of false positives and enhance positioning accuracy, we refrain from annotating distant poles, even if it may introduce the possibility of false negatives.

\subsubsection{Segmentation-based annotation (denoted S)} 
We use the HRNet image semantic segmentation neural network proposed in~\cite{hrnet} and pre-trained on the BDD100K dataset~\cite{bdd100k} to extract pole bases from the segmentation masks as in~\cite{ITSC23}.
We combine all pole-related classes to form entire pole clusters to check if they are connected to ground pixels. It ensures that only large clusters of pole pixels are considered, thus minimizing the influence of poor segmentation. However, some poles can be merged during clustering. To avoid this, we have chosen to extract any small clusters of pixels lying on the ground.
An example of segmentation-based annotations is visible in Fig. \ref{fig:segmentation_annotation} with the corresponding segmentation mask. 
Note that two of the poles missed by the map-based annotation are now annotated, even if one is still missing. 
However, some wrong annotations were introduced in the process: some correspond to bollards and some to other errors.

\subsubsection{LiDAR-based annotation (denoted L)} We segment the point cloud using Cylinder3D, a 3D convolution network proposed in \cite{zhu2020cylindrical} and pre-trained on SemanticKITTI~\cite{semantic_kitti}. 
Then, we group the points classified as poles into clusters to identify each pole individually and fit a 3D bounding box.
The pole base 3D coordinates corresponds to the center of the bottom face of the bounding box, which is in turn projected on the image to generate the annotation.
An example of lidar-based annotations is visible in Fig. \ref{fig:lidar_annotation} along with clusters bounding boxes projection.
Similarly to the image segmentation annotation, the neural network trained for the lidar segmentation consider that bollards belong semantically to the pole class.
In this particular example, some bollards were detected and some others not. Far poles were not annotated due to the data sparsity and the lack of points.


\subsection{Annotations association and fusion}
\label{sec:annotations_fusion}

For the association of the annotations, we use a simple unique nearest neighbor approach with the Euclidean distance to associate the annotations between two sets.
Two annotations across two modalities are grouped if their relative distance is smaller than a given threshold.
If an annotation can be associated with more than one, only the closest one is kept.
The final association function \(h\) with three modalities consists in grouping all the pairwise associations.

Once a set \(A\) of annotations is computed, the fused annotation \(f(A)\) is defined as the best annotation contained in \(A\) considering a preference order: \(S\succ L\succ M\). This order is based on the presence of potential sources of positioning errors in the image provided by the different automatic methods. Only segmentation errors can occur from the method \(S\). Errors due to sensor calibration, segmentation and clustering can impact the method \(L\). Finally, errors due to vehicle positioning errors, sensor calibration, map and ground segmentation can arise from the method \(M\). 
This preference order is further justified in the experiment results in Sec.~\ref{sec:single_annotation}.

\subsection{Ambiguous annotations management}

\label{sec:detector}

As described in Sec.~\ref{sec:consensus}, different fused annotation sets can be generated depending on a degree of consensus 
\(q\).
We may wish to train a detector only on annotated examples with a high degree of consensus while disregarding those with low consensus that may not represent pole bases.

We establish two sets: \(A^*\) comprising automatic annotations with high consensus, serving as the labels for our training set, and \(\widetilde{A}\) containing  all other ambiguous automatic annotations annotated by at least one method that we aim to exclude. 
Given a minimum consensus threshold \(Q\), we have
\begin{equation}
    A^* = {}^{(1:K)}_{\phantom{111}Q} a \quad\text{and}\quad
    \widetilde{A} = \left.{}^{(1:K)}_{\phantom{1111}1} a \middle\backslash {}^{(1:K)}_{\phantom{111}Q} a\right. 
\end{equation}

Handling the ambiguous annotations \(\widetilde{A}\) is not straightforward. 
Adding the ambiguous annotations may lead to false positive labels while removing them may lead to false negative labels.
In both cases, these potentially erroneous labels may lead to a decrease of performance in the training.
In this paper, we propose a simple bypass by adding black squared patches to mask ambiguous annotations.

\begin{figure}
    \centering
    \includegraphics[width=\columnwidth]{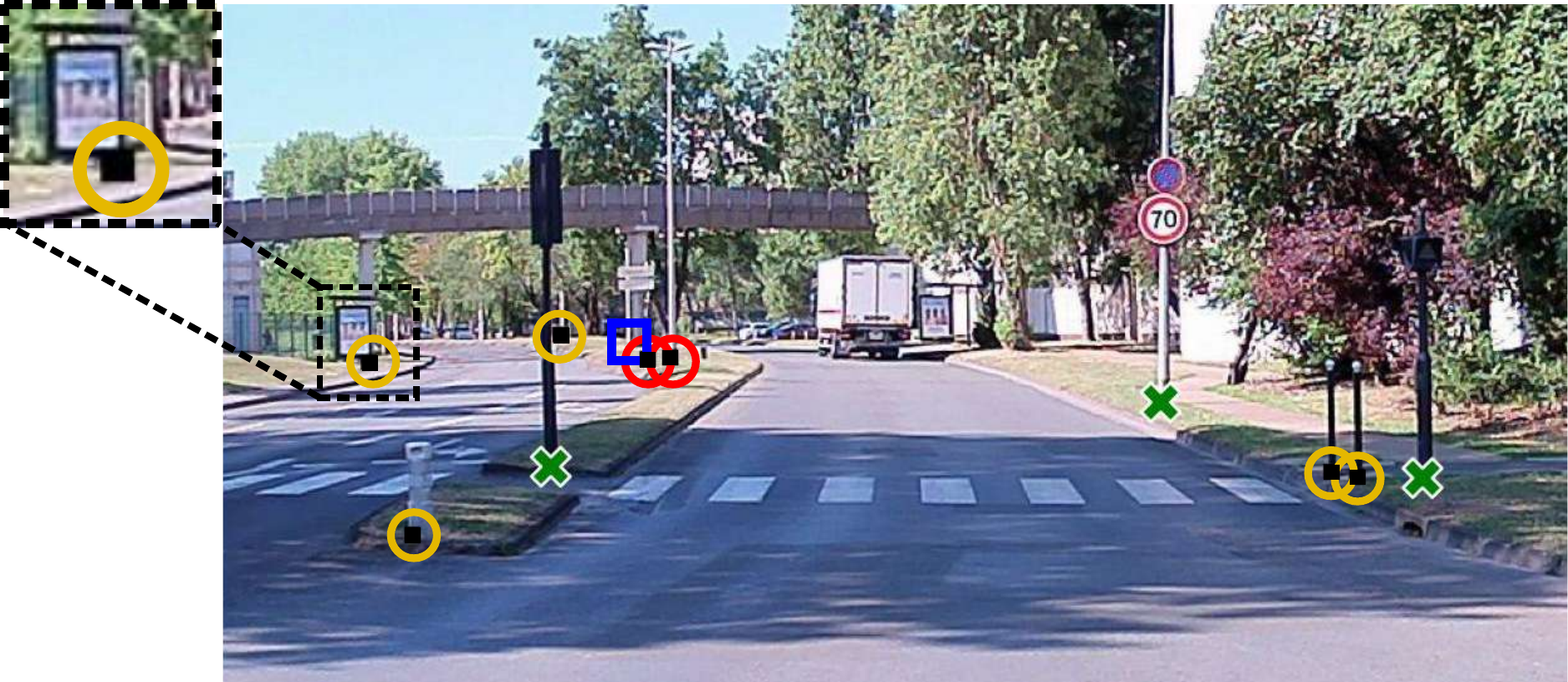}
    \caption{Management of ambiguous pole bases. Green crosses: annotations with unanimous agreement. 
    Orange circles: unnecessary black patches. 
    Red circles: black patches to mask ambiguous pole bases. 
    Blue square: missed pole base.}
    \label{fig:patches}
\end{figure}

Fig.~\ref{fig:patches} illustrates such kinds of instances. 
In this particular case, we set $Q=3$, \textit{i.e.}, a pole is annotated if all three modalities have annotated it as such (they are depicted with green crosses). 
For ambiguous annotation cases, a black patch can be added to mask a part of the image corresponding to two possible situations: i) One that does not correspond to a pole base (yellow circles in Fig.~\ref{fig:patches}): while unnecessary, it is not expected to impact the model training and ii) one that actually corresponds to a pole base not included in the final label set (red circles in Fig.~\ref{fig:patches}). Adding the patch is essential to help the model during training because otherwise it leads to a false negative annotation.
Even using multiple annotation methods, pole bases may still be missed, resulting in false negatives in the training set as seen with the blue square. To minimize the occurrence of such cases, the union of all methods must be able of annotating as many pole bases as possible. Simultaneously, a consensus must be established among the methods with a sufficient number of labels for training.

\section{Experimental results}
\label{sec:exp}

\subsection{Dataset}

To the best of our knowledge, there is no public dataset including an HD map with georeferenced poles along with image and lidar data. 

We conducted experiments using an experimental Renault ZOE equipped with the following sensors:
\begin{itemize}
    \item Hesai Pandora which integrates a 40-layer LiDAR with monocular cameras.
    \item NovAtel SPAN-CPT GNSS/IMU with PPK computations for high accuracy localization.  
\end{itemize}
We carried out five acquisitions, each lasting approximately one hour under different traffic conditions in 2022. 
For the map-based annotation of images, we used a 2D HD map of the city of Compi\`egne, France. 

Among these sequences, we selected two with different traffic and lighting conditions to manually annotate a subset of images for annotation and detection evaluation. Given our deliberate exclusion of short-lived objects such as bollards, we focused our annotations on elements as traffic signs, traffic lights, or streetlamps. Consequently, our objective is to develop a specialized detector for the elements stored in the HD map, which will be used for localization.

We extracted 939 images particularly representative of the sequences acquired and manually annotated 2846 poles\footnote{Dataset available on https://datasets.hds.utc.fr/share/vBMPrMDM2wOEd2g}. From the remaining sequences 5391 images were extracted and automatically annotated for subsequent training purposes.

\subsection{Single annotation methods}
\label{sec:single_annotation}

We evaluated the automatic annotation methods in terms of precision and recall. The results are reported in Tab.~\ref{tab:simple-annotations-metrics}.
S generates approximately five times as many pole-base candidates as the other two methods. This method is the most generic, annotating anything considered as a pole by the semantic segmentation network, resulting in the highest recall. However, its precision is much lower than \(M\) due to its generality. \(M\), although annotating fewer pole bases, achieves the best precision since our manual annotations correspond to the classes contained in the map. 
Because the map-based method relies on the lidar data, the recall gets limited by the range of the sensor and the sparsity of the point cloud.
\(L\)~exhibits a higher precision than \(S\), indicating fewer point cloud segmentation errors compared to image segmentation errors in our case.
Moreover, it is simpler to extract poles from segmented point clouds than from segmented images. 
However, similarly to \(M\) the recall of \(L\) is lower due to the lidar sensor limitation.

\begin{table}
\caption{Annotation evaluation of the three basic methods.
Number: number of annotated poles; FP: false positive; TP: true positive; FN: false negative; Prec: precision (\%); Rec: recall (\%); MAE-x: median absolute error in pixels along the x-axis.
} 
\label{tab:simple-annotations-metrics} 
\smallskip
\footnotesize{
\begin{tabularx}{\columnwidth}{cccccccc}
 \toprule
 Method  & Number  & FP & TP  & FN & Prec. & Rec.  & MAE-x \\
 \cmidrule(lr){1-8}
    \(M\) & 1364 & \textbf{232} & 1132 & 1714 & \textbf{83.0} & 39.8 & 4.34\\
   \(S\)  & 6187 & 3757 & \textbf{2430} & \textbf{416} & 39.3 & \textbf{85.4} & \textbf{1.06} \\
    \(L\) & 1231 & 492 & 739 & 2107 & 60.0 & 26.0 & 2.68\\
 \bottomrule
\end{tabularx}
}
\end{table}

To accurately detect pole bases, our goal is to minimize horizontal positioning errors in the image frame. In fact, we suppose that errors along the y-coordinate are less impactful, given the vertical nature of pole objects and the use of the detection in a localization context. 
The median absolute error in Tab.~\ref{tab:simple-annotations-metrics} shows that \(S\) has a pixel-level error, \(L\) has an error twice as large and the error for \(M\) is four times higher.

This confirms the preference order chosen in Sec.~\ref{sec:annotations_fusion} in terms of positioning accuracy: \(S\succ L\succ M\).

\subsection{Annotation combinations}

We assessed various fusion strategies as previously defined, testing unions and intersections of sets. To build our pole base detector aligned with the HD map, we specifically consider combinations involving \(M\) as our objective is to detect elements present in the map for localization. We exclude combinations obtained only from \(L\) and \(S\), as they may annotate pole structures as bollards, which are not accounted for in the map.
The results are presented in Table~\ref{tab:fusion-annotations-metrics}.

\begin{table}
\caption{Annotation evaluation of the possible fusion strategies. ``\(|\)'' and ``\(\&\)'' indicate respectively union and intersection of annotations.  
}

\label{tab:fusion-annotations-metrics} 
\smallskip
\begin{tabularx}{\columnwidth}{ccccccc}
 \toprule
 Method  & Number  & FP & TP  & FN & Prec. & Rec.  \\
 \midrule
    \(M\) $|$ \(L\) & 2035 & 719 & 1316 & 1530 & 64.7 & 46.2 \\ 
    \(M\) $|$ \(S\) & 6424& 3917 & 2507 & 339 & 39.0 & 88.1 \\
    \(M\) $|$ \(S\) $|$ \(L\) & 6695 & 4169 & \textbf{2526} & \textbf{320} & 37.7 & \textbf{88.7} \\
 \midrule
    \(M\) \& \(L\) & 560 & 24 & 536  & 2310 & 95.7 & 18.8 \\
    \(M\) \& \(S\) & 1127 & 84 & 1043 &1803 & 92.5 & 36.6 \\
    \(M\) \& \(S\) \& \(L\) & 513 & \textbf{5} & 508 & 2338 & \textbf{99.0} & 17.8\\
 \bottomrule
\end{tabularx}
\end{table}

As expected, the union of annotation sets improves the recall. The recall is  limited (less than \(50\%\)) when \(S\) is not involved. However, \(S\) decreases strongly the precision of the union by almost one half.

Conversely, applying the intersection significantly improves precision to more than \(90\%\). 
Here, despite the poor \(S\) results in terms of precision (less than \(40\%\)), it does not negatively impact the precision of intersections. However, the intersection drastically reduces the number of annotations and the recall, even if the best precision is obtained when involving all methods for a small recall. \(M\) \& \(S\) is the best intersection of sets possible to guarantee the highest recall possible with a high precision.

Combinations involving at least \(M\) and \(S\) seem to be the best candidates for training. The union of at least \(M\) and \(S\) provides a very high recall, indicating that few pole bases are missed. On the other hand, the intersection ensures very high precision, thus limiting false candidates. It therefore seems possible to create new images using these approaches following the method described in Sec. \ref{sec:detector}. 

As a conclusion, the recall is maximized by \(M\) $|$ \(S\) $|$ \(L\), while the precision is maximized by \(M\) \& \(S\) \& \(L\) and the method \(M\) \& \(S\) provides an interesting compromise.
We focus on these three types of fusion in the following stage.

\subsection{Neural network training without ambiguity handling}

We evaluated the quality of automatic annotation for the training of a pole detector.
We formulated the detection problem as a standard object detection one by transforming a pointwise label into a squared \(250\times250\) pixels bounding box centered on the annotation~\cite{IV23}.
We used the YOLOv7~\cite{yolov7} model as the object detector.
We first constructed a reference model by training the YOLO model using the pole bases extracted from BDD100K~\cite{bdd100k} as in~\cite{IV23}.
It was then fine-tuned using our automatically annotated images using different approaches.
All the models were trained on a single Tesla V100 32G GPU for 300 epochs.

\begin{figure}
    \centering
    \includegraphics[width=\columnwidth]{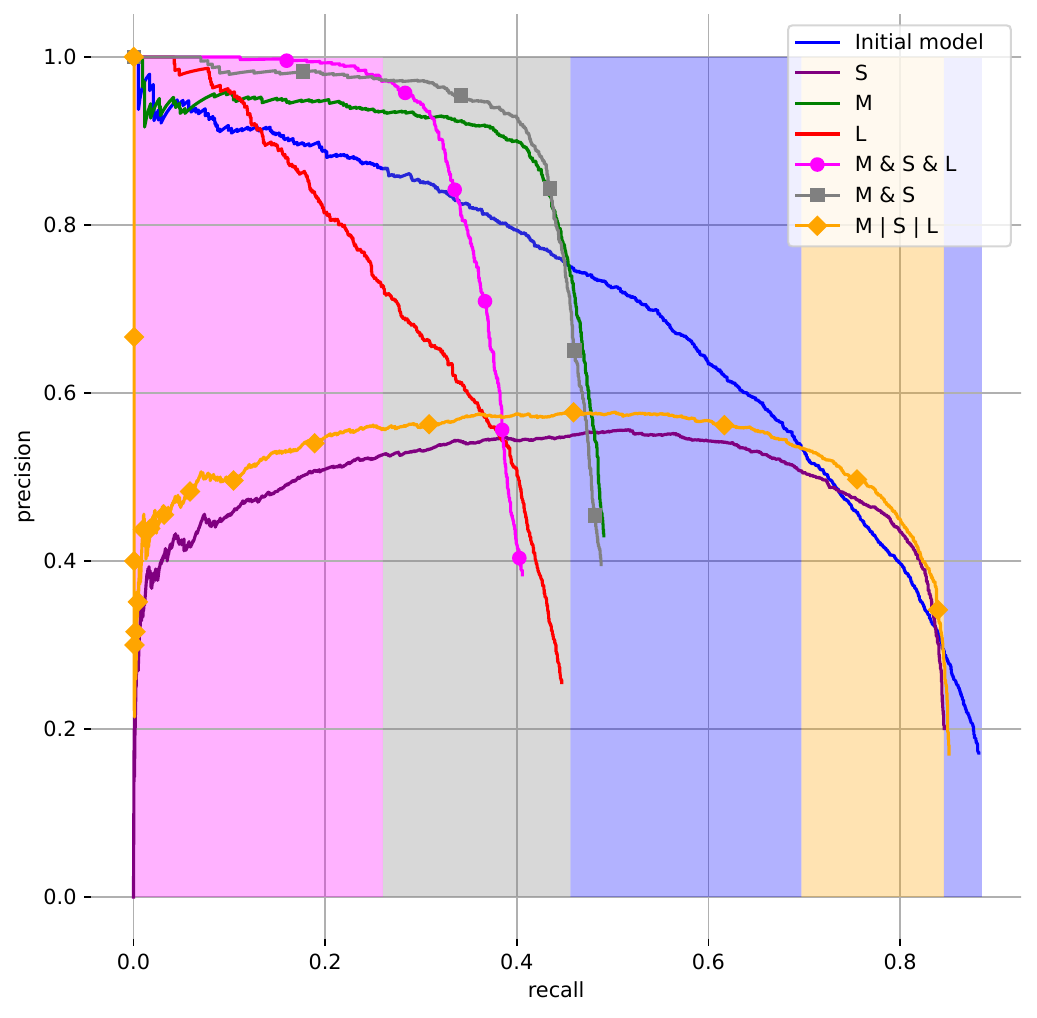}
    \caption{Precision-recall curves after 300 epochs of training using different annotation approaches. The background color indicates the predominant curve.}
    \label{fig:pr-curve}
\end{figure}

To visualize the impact of the training, we tested the final model obtained at the last epoch on our set of manually annotated images. The detection results obtained for all annotation methods and many combinations are summarized in Fig.~\ref{fig:pr-curve} by precision-recall (PR) curves following a standard object detection evaluation, \textit{i.e.}, using the IoU criteria between the detected and ground truth bounding boxes (generated from the pointwise labels). Results of the initial model are also indicated in blue.

Firstly, the results obtained with the methods \(M\) and \(L\) align with previously obtained annotation results. \(M\) achieves noteworthy precision, surpassing our initial model. However, it faces challenges in achieving high recall, even at the cost of sacrificing precision. \(L\) yields similar results, with a larger drop of precision when increasing recall and a smaller maximal possible recall. 

The application of the \(S\) method for training, due to its generality, struggles to attain a precision above approximately 0.58. At low recall values, indicating high confidence thresholds, the precision is notably low. This is attributed to objects with high confidence that are not considered as pole bases in our study, such as bollards. The curve rises upon including the other detections with lower confidence, yet corresponding to true positives.
Overall, \(S\) achieves much higher recall than \(M\) but with a much lower precision.

In Fig.~\ref{fig:pr-curve}, for different ranges of recall, the background color indicates the method with the highest precision.
We can see that, even though all single methods yield different results, none are optimal compared to combinations or the baseline model. No single method dominates any part of the graph.

\(M\) \& \(S\) \& \(L\) improves the performance in terms of precision, dominating all models in terms of precision, for recall objectives below approximately 0.25. However, the PR curve falls below the \(M\) \& \(S\) curve for higher recall objectives,  despite having higher precision of annotation.  This likely results from intersecting LL with any other method, which significantly reduces the number of usable annotations, thereby reducing recall performance. Removing \(L\) substantially improves the recall while still keeping a high precision. 
\(M\)  \& \(S\)  dominates all models for a recall ranged from 0.25 to 0.45 approximately. The precision is still close to \(M\) \& \(S\) \& \(L\) for recalls below 0.25. 
However, for higher recall objectives, the baseline model outperforms \(M\) \& \(S\) illustrating the limitation of the intersection of sets to reach high recall with sufficient precision.

The union of all annotations has a similar impact to \(S\) due to the presence of all \(S\) annotations and is less interesting than other combinations. Yet, for recall objectives between around 0.7 and 0.85, it surpasses the baseline. 

\subsection{Training with masking patches}

\begin{figure}
    \centering
    \includegraphics[width=\columnwidth]{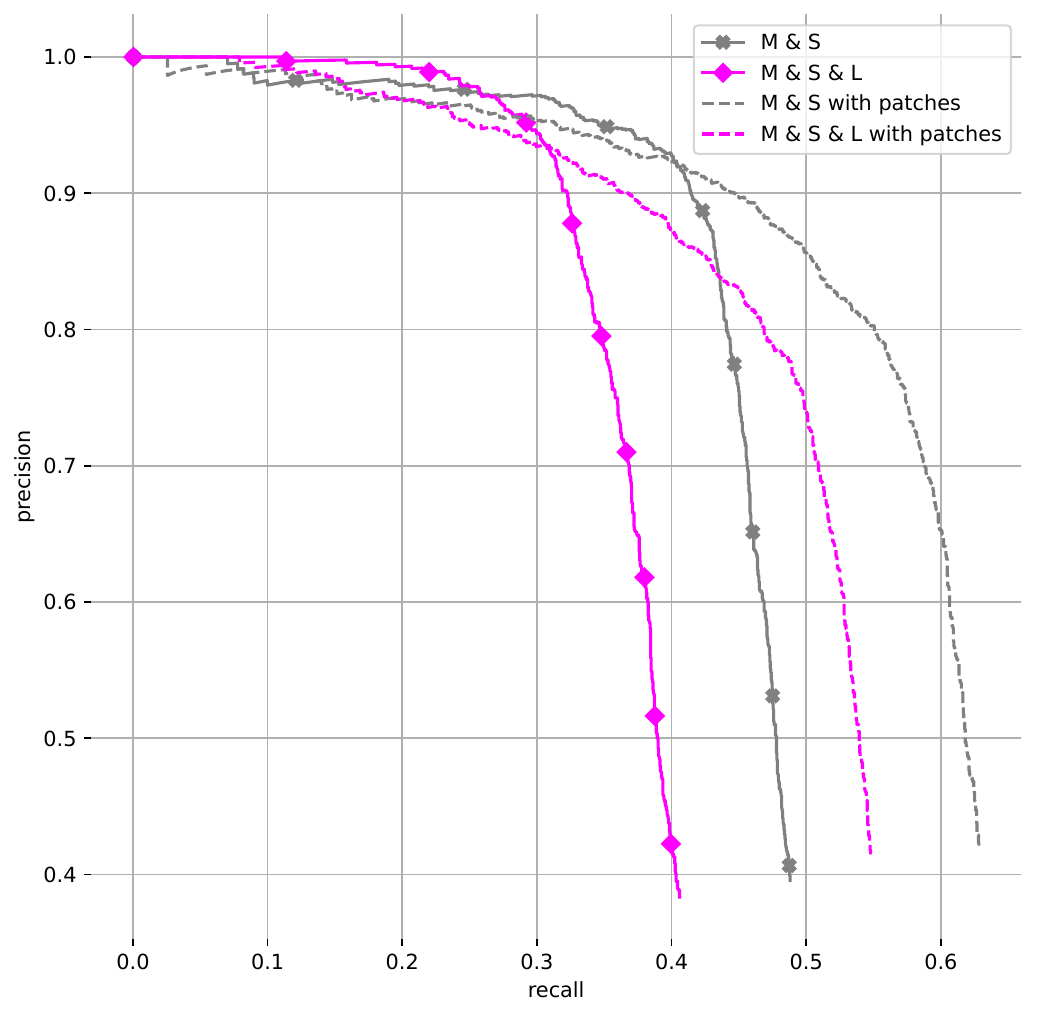}
    \caption{Precision-recall curves after 300 epochs of training using black patches on images.}
    \label{fig:pr-curve-masks}
\end{figure}

In the results presented in the previous section, the ambiguous annotations, that may occur in the \(M\) \(\&\) \(S\) and \(M\) \(\&\) \(S\) \(\&\) \(L\) setups, were simply discarded.
Fig.~\ref{fig:pr-curve-masks} shows the results obtained when adding the black patches as described in Sec.~\ref{sec:detector}.
A significant gain in terms of recall is observed, more than \(10\%\) with slight decrease in precision within the low ranges of recalls.
Generally speaking, adding the black patches to mask the ambiguous labels is clearly beneficial for the training of the object detector.





\section{Conclusion}

In this paper, we introduced a framework to combine different automatic methods to annotate unlabeled images from multi-modal raw data in order to train a pole base detector associated with features encoded in an HD map.
We proposed a way to manage ambiguous labels by masking parts of images with patches which  improved performance on a dataset with map data specifically collected for this study. 
The different manners to combine the individual annotation sources led to various precision and recall behaviors. 

In future work, we will study the choice of a relevant compromise between these two properties. This depends on the subsequent localization system according to its robustness (ability to reject outliers) and its need for exteroceptive information to compute accurate and reliable poses.

\label{sec:conclusion}

\bibliographystyle{IEEEtran}
\bibliography{biblio}
\end{document}